\documentclass[twocolumn]{style}

\usepackage{booktabs}

\newcommand{\ua}{\uparrow}
\newcommand{\nc}{\newcommand}
\nc{\da}{\downarrow} \nc{\hc}{\hat{c}} \nc{\hS}{\hat{S}}
\nc{\bra}{\langle} \nc{\ket}{\rangle} \nc{\eq}{equation (\ref}
\nc{\h}{\hat} \nc{\hT}{\h{T}}\nc{\be}{\begin{eqnarray}}
\nc{\ee}{\end{eqnarray}}\nc{\rd}{\textrm{d}}\nc{\e}{eqnarray}\nc{\hR}{\hat{R}}\nc{\Tr}{\mathrm{Tr}}
\nc{\tS}{\tilde{S}}\nc{\tr}{\mathrm{tr}}\nc{\8}{\infty}\nc{\lgs}{\bra\ua,\phi|}\nc{\rgs}{|\ua,\phi\ket}
\nc{\hU}{\hat{U}}\nc{\lfs}{\bra\phi|}\nc{\rfs}{|\phi\ket}\nc{\hZ}{\hat{Z}}\nc{\hd}{\hat{d}}\nc{\mD}{\mathcal{D}}
\nc{\bd}{\bar{d}}\nc{\bc}{\bar{c}}\nc{\mc}{\mathcal}\nc{\ea}{eqnarray}\nc{\mG}{\mathcal{G}}\nc{\bce}{\begin{center}}
\nc{\ece}{\end{center}}
\date{12th December 2011}

\begin{document}

\title{Is Hamming distance the only way for matching binary image feature descriptors?}

\author{E. Bostanci}

\abstract{Brute force matching of binary image feature descriptors is conventionally performed using the Hamming distance. This paper assesses the use of alternative metrics in order to see whether they can produce feature correspondences that yield more accurate homography matrices. Two statistical tests, namely ANOVA (Analysis of Variance) and McNemar's test were employed for evaluation. Results show that Jackard-Needham and Dice metrics can display better performance for some descriptors. Yet, these performance differences were not found to be statistically significant.}

\maketitle

\section{Introduction}
\label{sec:introduction}
Binary image feature descriptors such as BRIEF~\cite{1}, ORB~\cite{2} and BRISK~\cite{3} are becoming quite popular due to their easiness in computation, simple structure and high performance especially for applications aiming to run at decent video rates~\cite{4}. Many vision applications require finding image correspondences for computing a homography matrix which models the perspective transformation between two images and represents a linear relationship between image features.

The process of calculating a homography involves finding matching features detected across two images and then using an established method, \emph{e.g.} RANSAC, to find the transformation supported by the majority of the feature correspondences. Previous work has shown that spatial distribution of these image features across the images play an important role in the accuracy of the calculated homography~\cite{5}. 

In practice, matching binary features is performed using two approaches. The first approach is known as `brute force' matching, conventionally done using the Hamming distance given that it can be implemented efficiently using \textit{XOR} instruction on bit sets, in which matching is performed by comparing every descriptor in the first image with the descriptors from the second image. The second approach, known as Fast Library for Approximate Nearest Neighbours (FLANN)~\cite{4}, makes use of either a randomized kd-tree or a hierarchical clustering tree, depending on the feature dataset, to optimize performance for speed. 

This paper investigates use of alternative approaches which are already used for comparing binary sequences in machine learning and pattern recognition~\cite{6} for the former matching method. A statistical evaluation will show the performance differences, in terms of the accuracy of the calculated homography matrix, when various distance metrics are employed. 

\section{Distance metrics}
\label{sec:metrics}
Considering two binary descriptors to be matched as two binary sequences, the following four dependent quantities are defined as:
\begin{equation}
\begin{array}{l}
\displaystyle f_{00}: \text{num. of positions where both descriptors have 0s.}\\
\displaystyle f_{01}: \text{num. of positions where the first has 0 and the second has a 1.}\\
\displaystyle f_{10}: \text{num. of positions where the first has 1 and the second has a 0.}\\
\displaystyle f_{11}: \text{num. of positions where both descriptors have 1s.}\\
\end{array} 
\label{eq:ff}
\end{equation}

Using these quantities, the following distance metrics used for evaluation are defined: $d_H$ for Hamming distance, $d_J$ for Jaccard-Needham, $d_C$ for correlation, $d_D$ for Dice and $d_Y$ for Yule distance metric. These are given by the literature~\cite{6, 7} as follows:
\begin{equation}
\begin{array}{l}
\displaystyle d_H = \frac{f_{11}+f_{00}}{f_{00}+f_{01}+f_{10}+f_{11}}\\
\displaystyle d_J = \frac{f_{10}+f_{01}}{f_{11} + f_{10} + f_{01}}\\
\displaystyle d_C = \frac{1}{2} -  \frac{f_{11}f_{00}-f_{10}f_{01}}{2\sigma}\\
\displaystyle d_D = \frac{f_{10} + f_{01}}{2f_{11} + f_{10} + f_{01}}\\
\displaystyle d_Y = \frac{f_{10}f_{01}}{f_{11}f_{00}+f_{10}f_{01}}
\end{array} 
\label{eq:formulas}
\end{equation}
where $\sigma = \sqrt{(f_{10}+f_{11})(f_{01}+f_{00})(f_{11}+f_{01})(f_{00}+f_{10})}$. These various metrics, widely used in data mining applications, are employed here in order to assess the accuracy of the homography matrix calculated using them.

\section{Evaluation}
\label{sec:evaluation}
The main question of this evaluation is first modelled in a \textit{null hypothesis} framework (two-way ANOVA with image pairs and distance metrics as the two independent variables). Here, the null hypothesis states that there are no differences between the accuracies of the homography matrices calculated using the correspondences found when various metrics are employed, given the same set of keypoints. The alternative hypothesis states that there will be differences if different metrics are used for matching these binary descriptors. Further analysis is performed using McNemar's test to identify pairwise differences.

\begin{figure}[h!t!p]
\centering{\includegraphics[width=0.75\columnwidth]{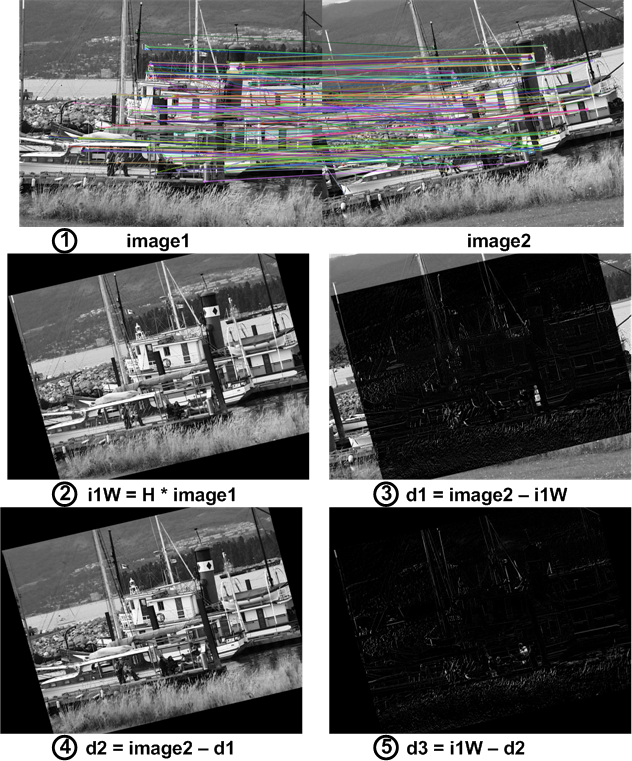}}
\caption{Process used in the evaluation}
\label{fig:explanation}
\end{figure}

In order to quantify the accuracy of the calculated homography matrices, this evaluation employed the following approach depicted in Fig.~\ref{fig:explanation}: First image features are extracted using the detector part of ORB (5000 keypoints were extracted in order to provide good coverage~\cite{5}). These keypoints, with computed descriptors (\textit{e.g.} BRIEF, BRISK), are matched across a pair of images. Using these matches, a homography matrix is calculated and applied to the first image in order to warp it onto the second. Here, an intermediate image is calculated ($d1$) which is then subtracted from the second image in order to remove the non-overlapping part ($d2$). Finally, the result image ($d3$) is obtained as the difference between warped version of the first image and $d2$. The evaluation criterion is chosen as the sum of non-zero pixels of $d3$. A larger sum indicates accuracy problems in the homography matrix resulting in alignment issues.

It is important to reiterate that the evaluation here focuses only on the selection of the distance metric for finding a correspondence, not for evaluating different descriptors. However, descriptors were varied in type and size (\emph{e.g.} BRIEF 32 and 64 bits) in order to see how this was affecting the results.

Using the sums of remaining pixels (samples for this evaluation) obtained from a dataset of 288 image pairs (all pairwise combinations) from a publicly available dataset\footnote{\url{http://www.robots.ox.ac.uk/~vgg/research/affine/}}, we applied a logarithmic transformation to the samples in order to reduce variance.

\section{Results}
\begin{figure}[h!t!p]
\centering{\includegraphics[width=.85\columnwidth]{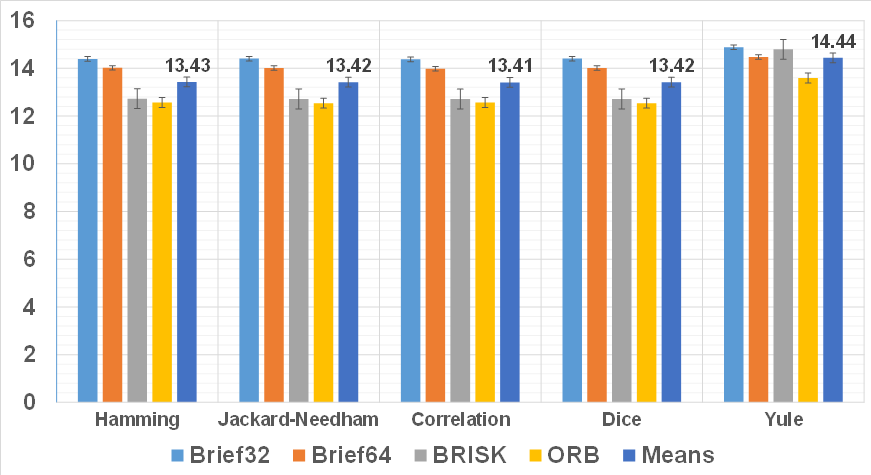}}
\caption{Effect of distance metric for different feature descriptors. Means bar indicate the mean performance of the metric in various descriptors.}
\label{fig:results}
\end{figure}

Initial results from the samples (Fig.~\ref{fig:results}) show some differences across the various metrics and descriptor types. In order to see whether these differences are statistically significant or not, ANOVA was employed (Table~\ref{tab:anova}). Analysis showed that different metrics have an effect on the results and the test confirms that there are significant differences ($F \gg F crit$, for all descriptor types). Furthermore, the \textit{P-value} shows that the confidence on these results is very high.
\begin{table}[h!t!b!p]
 \processtable{Results for the ANOVA test}
{    \begin{tabular}{llllll}
    \toprule
    \multicolumn{1}{c}{\textit{Source of Variation}} & \multicolumn{1}{c}{\textit{df}} & \multicolumn{1}{c}{\textit{MS}} & \multicolumn{1}{c}{\textit{F}} & \multicolumn{1}{c}{\textit{P-value}} & \multicolumn{1}{c}{\textit{F crit}} \\
    \midrule
    \multicolumn{6}{c}{\textbf{BRIEF32}} \\
    Image pairs & 287   & 26.26 & 88.83 & 0.00  & 1.16 \\
    Metrics & 4     & 13.75 & 46.53 & 0.00  & 2.38 \\
    Error & 1148  & 0.30  &       &       &  \\
    \midrule
    \multicolumn{6}{c}{\textbf{BRIEF64}} \\
    Image pairs & 287   & 42.51 & 93.07 & 0.00  & 1.16 \\
    Metrics & 4     & 12.52 & 27.41 & 0.00  & 2.38 \\
    Error & 1148  & 0.46  &       &       &  \\
    \midrule
    \multicolumn{6}{c}{\textbf{BRISK}} \\
    Image pairs & 287   & 134.53 & 63.60 & 0.00  & 1.16 \\
    Metrics & 4     & 248.40 & 117.44 & 0.00  & 2.38 \\
    Error & 1148  & 2.12  &       &       &  \\
    \midrule
    \multicolumn{6}{c}{\textbf{ORB}} \\
    Image pairs & 287   & 154.47 & 128.49 & 0.00  & 1.16 \\
    Metrics & 4     & 62.36 & 51.87 & 0.00  & 2.38 \\
    Error & 1148  & 1.20  &       &       &  \\
    \bottomrule
    \end{tabular}%
    \label{tab:anova}
}{}%
\end{table}%

Having found the differences, the next step is to identify the distance metric which is producing better results in terms of the accuracy of calculated homography matrix. McNemar's test~\cite{8} was employed for this purpose allowing pairwise comparisons between different metrics using z-scores. A value of $0$ indicates no difference in performance, while $2.576$ indicates $99.5\%$ confidence in one-tailed prediction meaning that one metric is statistically better than the other~\cite{8}. Looking at the results of Table~\ref{tab:mcnemar}, it can be seen that Yule displayed the worst performance of all the metrics for different descriptors.  For BRISK and ORB descriptors, Jaccard-Needham has surpassed the Hamming distance; however, the difference was not statistically significant. Comparing this result with that of Fig.~\ref{fig:results}, on average, there are metrics that perform better than the Hamming distance, but this did not manifest itself significantly in a statistical evaluation.
\begin{table}[h!t!b!p]
 \processtable{Results for the McNemar's test. Arrow-heads point to the metric resulting in a better accuracy.}
 {\begin{tabular}{lllll}
     \toprule
     \textit{Metric}      & Jaccard-Needham & Correlation & Dice  & Yule \\
     \midrule
     \multicolumn{5}{c}{\textbf{BRIEF32}} \\
     \midrule
     
     Hamming & $\leftarrow$ 0.84  & $\leftarrow$ 0.71  & $\leftarrow$ 0.84  & $\leftarrow$ 7.60 \\
     Jaccard-Needham &       & $\leftarrow$ 0.07  & 0.00  & $\leftarrow$ 7.72 \\
     Correlation &       &       & $\uparrow$ 0.07  & $\leftarrow$ 7.72 \\
     Dice  &       &       &       & $\leftarrow$ 7.72 \\
     \midrule
     \multicolumn{5}{c}{\textbf{BRIEF64}} \\
     Hamming & $\leftarrow$ 1.30  & $\leftarrow$ 0.45  & $\leftarrow$ 1.30  & $\leftarrow$ 4.30 \\
     Jaccard-Needham &       & 0.00  & 0.00  & $\leftarrow$ 4.42 \\
     Correlation &       &       & 0.00  & $\leftarrow$ 4.72 \\
     Dice  &       &       &       & $\leftarrow$ 4.42 \\
     \midrule
     \multicolumn{5}{c}{\textbf{BRISK}} \\
     Hamming & $\uparrow$ 0.71  & $\leftarrow$ 0.97  & $\uparrow$ 0.71  & $\leftarrow$ 9.40 \\
     Jaccard-Needham &       & $\leftarrow$ 1.11  & 0.00  & $\leftarrow$ 10.70 \\
     Correlation &       &       & $\uparrow$ 1.11  & $\leftarrow$ 10.11 \\
     Dice  &       &       &       & $\leftarrow$ 10.70 \\
     \midrule
     \multicolumn{5}{c}{\textbf{ORB}} \\
     Hamming & $\uparrow$ 0.52  & $\leftarrow$ 1.56  & $\uparrow$ 0.52  & $\leftarrow$ 5.86 \\
     Jaccard-Needham &       & $\leftarrow$ 0.13  & 0.00  & $\leftarrow$ 6.34 \\
     Correlation &       &       & $\uparrow$ 0.13  & $\leftarrow$ 5.98 \\
     Dice  &       &       &       & $\leftarrow$ 6.34 \\
     \bottomrule
     \end{tabular}%
     \label{tab:mcnemar}
    }{}

\end{table}%

\section{Conclusion}
This paper presented an evaluation in order to see whether distance metrics other than the conventional Hamming distance can be used for matching binary image feature descriptors. Results revealed that there are, indeed, metrics resulting in a more accurate homography matrix, producing less difference when the first image is warped onto the second. Findings also showed that these differences did not reflect well in a statistical evaluation, \emph{i.e.} no significant differences were found. Considering the options provided by modern processor instruction sets for implementing the Hamming distance, it keeps its position to be the metric of choice for brute force matching of binary descriptors. 

\vskip5pt

\noindent E. Bostanci (\textit{Computer Engineering Department, \\Ankara University, Turkey})
\vskip3pt

\noindent E-mail: ebostanci@ankara.edu.tr


\begin{thebibliography}{}

\bibitem{1}
Calonder, M., Lepetit, V., Ozuysal, M., Trzcinski, T., Strecha, C. and Fua, P.: `BRIEF: Computing a Local Binary Descriptor Very Fast', \textit{IEEE Transactions on Pattern Analysis and Machine Intelligence}, 2012, \textbf{34}, (7) pp. 1281--1298

\bibitem{2}
Rublee, E. and Rabaud, V. and Konolige, K. and Bradski, G.: `ORB: An efficient alternative to SIFT or SURF', \textit{International Conference on Computer Vision}, 2011, pp. 2564-2571

\bibitem{3}
Leutenegger, S., Chli, M. and Siegwart, R.Y.: `BRISK: Binary Robust invariant scalable keypoints', \textit{International Conference on Computer Vision}, 2011, pp. 2548-2555

\bibitem{4}
Muja, M. and Lowe, D. G.: `Fast Matching of Binary Features', \textit{Ninth Conference on Computer and Robot Vision}, 2012,  pp. 404--410

\bibitem{5}
Bostanci, E., Kanwal, N. and Clark, A. F.: `Spatial Statistics of Image Features for Performance Comparison', \textit{IEEE Transactions on Image Processing}, 2014, \textbf{23}, (1), pp. 153--162

\bibitem{6}
Choi, S. S.,  Cha, S. H. and Tappert, C.: `A Survey of Binary Similarity and Distance Measures', \textit{Journal on Systemics, Cybernetics and Informatics}, 2010, \textbf{8}, (81), pp. 43--48

\bibitem{7}
Warrens, M. J.: `Similarity Coefficients for Binary Data', \textit{PhD Thesis}, Universiteit Leiden, 2008

\bibitem{8}
Bostanci, B. and Bostanci, E.: `An Evaluation of Classification Algorithms Using Mc Nemar's Test', \textit{Proceedings of Seventh International Conference on Bio-Inspired Computing: Theories and Applications}, 2013, \textbf{201}, pp. 15--26

\end{thebibliography}
\end{document}